\pdfoutput=1
\documentclass[11pt]{article}

\usepackage[]{ACL2023}

\usepackage{lipsum}
\usepackage{arydshln}

\usepackage{times}
\usepackage{latexsym}
\usepackage{hyperref}
\usepackage{comment}
\usepackage[T1]{fontenc}
\usepackage[utf8]{inputenc}

\usepackage{microtype}

\usepackage{inconsolata}

\usepackage{times}
\usepackage{latexsym}

\usepackage[T1]{fontenc}
\usepackage[utf8]{inputenc}

\usepackage{microtype}

\usepackage{inconsolata}

\usepackage{footmisc}
\usepackage{graphicx}
\usepackage{multirow}
\usepackage{tabularx}
\usepackage{csquotes}
\usepackage{booktabs}
\usepackage{bm}
\usepackage{CJKutf8}
\usepackage{array}
\usepackage{latexsym}
\usepackage{amsthm,amsmath,amssymb}
\usepackage{array}
\usepackage{url}
\usepackage{float}

\usepackage{xcolor}
\usepackage{enumitem}
\usepackage[normalem]{ulem}

\usepackage{tabularx}
\usepackage{booktabs}
\usepackage{multirow}
\usepackage{graphicx}
\usepackage{amsmath}
\usepackage{enumitem}
\usepackage{adjustbox}

\usepackage{array}
\newcolumntype{L}[1]{>{\raggedright\arraybackslash}p{#1}}

\setlist[itemize]{itemsep=0.02cm,topsep=0.2cm}

\usepackage{xcolor}
\usepackage[normalem]{ulem}

\def\ZKdel#1{\bgroup\markoverwith{\textcolor{green!60!black!100}{\rule[0.4ex]{2pt}{3pt}}}\ULon{#1}}

\def\SMdel#1{\bgroup\markoverwith{\textcolor{red!60!black!100}{\rule[0.4ex]{2pt}{3pt}}}\ULon{#1}}
\def\OD#1{{\color{cyan!80!yellow!80!black!100}OD: \it #1}}
\def\ODdel#1{\bgroup\markoverwith{\textcolor{cyan!80!yellow!80!black!100}{\rule[0.4ex]{2pt}{3pt}}}\ULon{#1}}

\def\MLdel#1{\bgroup\markoverwith{\textcolor{blue!60!black!100}{\rule[0.4ex]{2pt}{3pt}}}\ULon{#1}}

\title{A Survey of Text Style Transfer: Applications and Ethical Implications}

\author{Sourabrata Mukherjee, Mateusz Lango, Zdeněk Kasner, Ondřej Dušek\\
Charles University, Faculty of Mathematics and Physics\\
Institute of Formal and Applied Linguistics\\
Prague, Czech Republic\\
\texttt{\{mukherjee,lango,kasner,odusek\}@ufal.mff.cuni.cz}}

\begin{document}
\maketitle
\begin{abstract}
% \MLdel{Text style transfer (TST) is an important approach in controllable text generation that aims to control certain attributes of language use, such as politeness, formality, and sentiment, without changing the style-independent content of the text. 
% It is a framework that has been increasingly gaining importance due to its applications in automatic text generation and its relevance for various language tasks. 
% Much work has been carried out in the TST domain recently, but not much has been discussed concerning its motivation factors and applications. This paper presents a comprehensive review of TST applications that have been the focus of research over the years using both traditional linguistic approaches and neural networks. We also discuss current challenges, future research directions, and the ethical implications of TST applications in text generation to stimulate further research. By providing a holistic overview of the landscape of TST applications, we hope to contribute to the advancement of research and understanding of the ethical considerations associated with TST.}
Text style transfer (TST) is an important task in controllable text generation, which aims to control selected attributes of language use, such as politeness, formality, or sentiment, without altering the style-independent content of the text. 
The field has received considerable research attention in recent years and has already been covered in several reviews, but the focus has mostly been on the development of new algorithms and learning from different types of data (supervised, unsupervised, out-of-domain, etc.) and not so much on the application side.
However, %with the advent of large language models, 
TST-related technologies are gradually reaching a production- and deployment-ready level, and therefore, the inclusion of the application perspective in TST research becomes crucial.
Similarly, the often overlooked ethical considerations of TST technology have become a pressing issue.\\
This paper presents a comprehensive review of TST applications that have been researched over the years, using both traditional linguistic approaches and more recent deep learning methods. We discuss current challenges, future research directions, and ethical implications of TST applications in text generation. By providing a holistic overview of the landscape of TST applications, we hope to stimulate further research and contribute to a better understanding of the potential as well as ethical considerations associated with TST.
\end{abstract}

\section{Introduction}

\begin{figure*}[t]
  \centering
  \includegraphics[width=.7\textwidth]{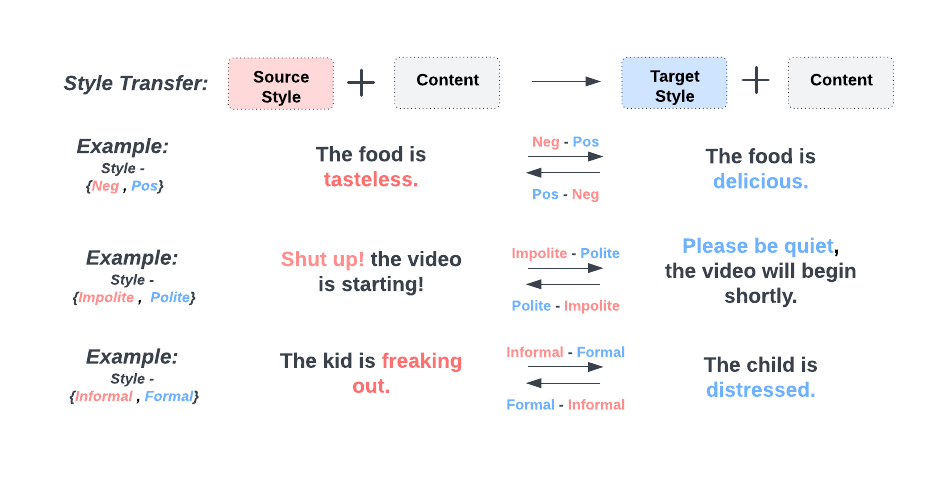}
  \caption{Several examples of text style transfer: transferring sentiment, politeness and formality.}
\label{fig:tst_examples}
\end{figure*}

Text Style Transfer (TST) is a task in controllable natural language generation that constructs techniques that change the stylistc attributes of a written text such as politeness, formality, or sentiment, without altering the rest of the text's meaning~\cite{jin-etal-2022-deep}.
Although the general goal of rewriting a text while preserving its content is relevant to other NLP tasks such as machine translation \cite{DBLP:conf/nips/SutskeverVL14, DBLP:journals/corr/BahdanauCB14} or summarisation~\cite{allahyari2017text}, TST is unique in focusing on  more subtle, stylistic aspects of the text. 
The text style may reflect the writer's demographic characteristics, such as gender and age, their educational background, emotional state, and even situational context, as reflected in the sentiment and politeness of the text.  %\ML{SM - why TST is unique? Can we charaterize TST better?}
Figure~\ref{fig:tst_examples} shows some basic examples of TST.

% Controllable text generation has multiple applications as outlined in \citet{DBLP:conf/eacl/McDonaldP85} and \citet{hovy1987generating}. These range from controlling persona in dialog response generation \cite{DBLP:conf/acl/KielaWZDUS18, DBLP:conf/acl/LiGBSGD16} to controlling various aspects of the response such as politeness \cite{niu2018polite}, grounding responses in external source of information \cite{DBLP:conf/emnlp/ZhouPB18, DBLP:conf/iclr/DinanRSFAW19, DBLP:conf/aaai/GhazvininejadBC18} and controlling the topic sequence \cite{DBLP:conf/acl/TangZXLXH19, DBLP:journals/corr/abs-2002-02878}. TST can also aid in other applications like story generation, in which it can control the ending \cite{peng2018towards}, the persona \cite{DBLP:journals/corr/abs-1906-06401}, the plot \cite{DBLP:conf/aaai/YaoPWK0Y19}, and even the topic sequence \cite{DBLP:conf/aaai/HuangGCWW019}.
  
In recent years, TST research has gained considerable momentum.
With the advent of neural networks, text style transfer research has shifted from traditional linguistic approaches, which relied on hand-crafted grammars and templates, to data-driven approaches, typically based on sequence-to-sequence modeling \cite{DBLP:conf/naacl/RaoT18}. However, since there is a considerable dearth of style-specific parallel data, TST research has taken up the challenge of devising new methodologies, such as latent-space content-style disentanglement \cite{hu2017toward,shen2017style}, prototype editing \cite{DBLP:conf/naacl/LiJHL18}, or pseudo-parallel corpora \cite{DBLP:journals/corr/abs-1808-07894,jin2019imat} (see Section~\ref{textstyletransfer} for details). 

% For instance, \citet{hu2017toward} and \citet{shen2017style} came up with the idea of disentangling text into its content and style attributes in the latent space and applying generative modeling. \citet{DBLP:conf/naacl/LiJHL18} follow prototype editing combining a sentence template and its attribute markers to generate text. Pseudo-parallel corpus construction for training the model in a supervised way with the pseudo-parallel data is yet another approach used by \citet{DBLP:journals/corr/abs-1808-07894} and \citet{jin2019imat}. These approaches were further refined with the emergence of the Transformer-based models by \citet{DBLP:conf/emnlp/SudhakarUM19},  \citet{DBLP:conf/emnlp/MalmiSR20}, \citet{mukherjee2022balancing}, and \citet{mukherjee-etal-2023-low}.

This rapid development of TST technologies has already been covered by several reviews~\cite{jin2022deep,hu2022text}, but they are method-oriented and do not pay much attention to the research on \textit{TST applications}, which has also come a long way.
We believe that this application perspective will play an increasingly important role, as
%, with the advent of large language models, 
TST-related technologies are gradually reaching a production- and deployment-ready level, which will boost interest in application-specific issues.
Therefore, in this paper, we provide what we believe to be the first comprehensive overview of different TST applications in order to stimulate future research in this area.
We analyze TST applications in user privacy and security, the generation of personalized texts, and building dialogue agents, as well as the use of TST to improve the performance of other NLP and machine learning tasks.
We also provide a discussion of the implications of TST applications from an ethical perspective, draw lines for future research, and review recent LLM-related TST work that has not been covered in previous reviews.

% \MLdel{The above examples showcase how far the field of TST has come. In \citet{jin2022deep} and \citet{hu2022text}, the authors report on the models and the methodologies for TST that have been researched over the years. Along with that, the research on \textit{TST applications} has also come a long way. However, to the best of our knowledge, there is no comprehensive survey covering this area of research. Thus, the main motivation behind this survey is to bring together and report on applications of TST and to provide directions for future work.}

We begin by defining the task of text style transfer and providing a brief summary of the models proposed in TST research in Section~\ref{textstyletransfer}. 
% We begin the survey by providing a definition of TST task and a brief summary of models proposed in TST research in Section~\ref{textstyletransfer}. 
This is followed by the main contribution of this paper, which is an overview of different applications of TST in Section~\ref{applications}.
% The key contribution of this paper is the overview of various applications of TST, which we discuss in Section~\ref{applications}. %: 
% \begin{itemize}
%   \item Section \ref{realworldapplications} portrays the real-world applications of TST, including fighting toxic and offensive language, controlling dialog response generation, simplification of the text, etc.
%   \item  Section \ref{subtasksofTST} categorises the sub-tasks of TST which can be extended into TST applications, including the task of changing the sentiment, formality, and politeness of the text, etc.
% \end{itemize}
Next, we describe open problems and future research directions in Sections~\ref{sec:openproblems} and~\ref{sec:future}.
 Finally, we conclude with a discussion of  the ethical impact of TST applications in Section~\ref{ethics}.

\section{The Task of Text Style Transfer} \label{textstyletransfer}

In this section, we present a basic introductory overview of TST. For more detail, we refer the reader to the existing general surveys of TST by \citet{jin2022deep} and \citet{hu2022text}. 

\textit{Text style} refers to how individuals express their state of mind, personal characteristics, and rhetorical choices in their writing. 
This can be seen in their word choice, writing style, and other attributes. 
%Text style can be conveyed through word choice, writing style, and other attributes. 
The style of a text can also be described using various concepts such as sentiment \cite{mukherjee2022balancing}, emotion \cite{helbig2020challenges}, humor \cite{hossain2019president}, similes \cite{chakrabarty2020generating}, personality \cite{kaptein2015personalizing}, politeness \cite{mukherjee2023polite}, formality \cite{DBLP:conf/naacl/RaoT18}, simplicity \cite{tan2017internet}, or authorship \cite{reddy2016obfuscating}.

TST models aim to change the style of the text while preserving the \textit{style-independent} content in the text. Given an arbitrary text, $x$ with a source style $s_1$, our goal is to rephrase $x$ to a new text $\hat{x}$ with a target style $s_2$ ($s_2\neq s_1$) while preserving its style-independent content (see Figure~\ref{fig:tst_examples}).

The training process of these models can be broadly classified into four categories which we discuss in the following subsections: parallel supervised (Section \ref{sec:par}), non-parallel supervised (Section \ref{sec:nonpar}), purely unsupervised (Section \ref{sec:unsupervised}), and using Large Language Models (LLMs) (Section \ref{sec:llms}). We also briefly discussed the evaluation process of the TST task in Section \ref{sec:tst_eval}.

% Since there is very little available aligned data in TST tasks, mostly non-parallel or unsupervised methodologies have been explored. However, in Section \ref{sec:par} we also discuss here the models which are supervised using parallel corpora. 

\subsection{Parallel Supervised Approaches}
\label{sec:par}

Parallel supervised TST models are trained using similar pairs of texts with similar content but different styles. The most common underlying architecture of these models is the sequence-to-sequence architecture, same as for for many other natural language generation tasks \cite{gatt2018survey}. Usually, a sequence-to-sequence model is trained on a parallel corpus wherein the text of the original style is fed into the encoder and the decoder outputs the corresponding text according to the target style. For example, \citet{jhamtani2017shakespearizing} used this approach to translate between modern and Shakespearean English (where ample training data exists), while %trained a sequence-to-sequence model with a pointer network on a parallel corpus and then applied the model to translate modern English phrases to Shakespearean English.
\citet{mukherjee2023leveraging} employed minimal parallel data and various low-resource methods for sentiment transfer.

Applying the sequence-to-sequence approach to this task is quite challenging due to the unavailability of parallel data \cite{hu2022text,mukherjee-etal-2023-low}. To handle this, data augmentation methods, such as pseudo-parallel datasets, have been explored  \cite{mukherjee2023leveraging,shang2019semi,jin2019imat,nikolov2018large,liao2018quase}. 
% Nevertheless, these approaches are limited by the lack of systematic evaluation of the generated pseudo-parallel datasets.\OD{this seems like a pretty strong claim, anything to back it up?} 

\subsection{Non-parallel Supervised Approaches}
\label{sec:nonpar}

An alternative method to using parallel datasets is the non-parallel supervised setting. 
In this setting, the TST models aim to transfer the style of texts without any knowledge of matching text pairs in different styles. Data for each individual target style is provided, but not aligned with other styles.

To train the TST models in the non-parallel supervised setting, researchers have proposed the strategy of disentangling the style and content in text \cite{gatys2015neural}.
There are three main types of style-content disentanglement strategies:

\begin{enumerate}[label={(\arabic*)}]
  \item \textbf{Explicit Style-Content Disentanglement} TST models for \textit{explicit style-content disentanglement} adopt a straightforward text replacement approach for generating texts of a target style. 
  %The advantage of this approach is its simplicity and explainability. 
  For example, \citet{DBLP:conf/naacl/LiJHL18} found that parts of a text that are associated with the original style can be replaced with new phrases associated with the target style. 
  %The text was then fed into a sequence-to-sequence model to generate a fluent text sequence in the target style. 
  % However, these approaches are not suitable for TST applications that go beyond simple phrase replacement such as changing the formality level of the text.
  \item \textbf{Implicit Style-Content Disentanglement} For \textit{implicit style-content disentanglement}, TST models first learn the latent representations of the content and the style of the given text. The latent representation of the original content is then combined with the latent representation of the desired target style to generate text in the target style. Techniques such as back-translation and adversarial learning \cite{shen2017style,zhao2018adversarially,fu2018style,prabhumoye2018style,hu2017toward} have been proposed to disentangle content and style in latent representations.

  % To improve the results of explicit and implicit style-content disentanglement methods, the technique of adversarial learning has been explored. Adversarial learning has been used to generate text output that is indistinguishable from real data \cite{shen2017style,logeswaran2018content,tian2018Structured,chen2018adversarial,zhao2018language,yin2019utilizing,park2019paraphrase} and to remove the style attributes in the latent representation of text \cite{zhao2018adversarially,fu2018style,yang2018unsupervised, lai2019multiple,john2019disentangled}.\OD{note that you mention adversarial learning in the previous paragraph, so this now feels like a repeat; also, are all the citations related to TST?}
  
  \item \textbf{No Style-Content Disentanglement} The most recent research has also explored performing TST without disentangling the text's style and content. These techniques include attribute-controllable generation, reinforcement learning, probabilistic modeling 
  % ,and a pseudo-parallel corpus 
  \cite{lampleSSDRB19,dai2019style,li2019domain,luo2019dual,He2020A}.
  % \OD{hang on, you put pseudo-parallel corpus in Sect 2.1, so it shouldn't be here anymore, no?}
\end{enumerate} 

% However, \citet{lai2019multiple} argued on the impracticality of adversarial training on the disentanglement of style and content.
% A limitation of the adversarial learning is also its dependency on a style classifier, which requires large annotated datasets and has limited accuracy.

\subsection{Unsupervised Approaches}
\label{sec:unsupervised}

% For both parallel and non-parallel supervised data settings, style labels were used to enable the supervised training of the TST models. However, in the purely unsupervised setting, only an unlabeled text corpus is available and the TST models need to be trained to perform style transfer without any knowledge of style labels.\OD{not sure how you define success in that case?} There are relatively few works that have explored this form of TST, mostly basing their approach on unsupervised representation learning \cite{radford2017learning,jain2019unsupervised,xu2020variational,shen2020educating}. Even though some of the results are encouraging, more studies need to be conducted to generalize purely unsupervised methods for other TST tasks.\OD{can you be more specific wrt the problems these systems have? this sounds too vague}

% Most of the TST studies discussed above assume that style-specific corpora (either parallel or non-parallel) are available. However, this section introduces TST models that perform style transfer in a purely unsupervised setting, where only a mixed corpus of texts with unspecified styles is available.

%There are relatively fewer studies on purely unsupervised TST \cite{radford2017learning,xu2020variational,shen2020educating}. 
Works on purely unsupervised TST are comparatively rarer.
An early study by \citet{radford2017learning} explored the properties of recurrent language models at the byte level. They trained an LSTM model on text processed as a sequence of UTF-8 encoded bytes in an unsupervised manner. Interestingly, they found a single neuron within the trained LSTM that directly corresponded to the sentiment. By manipulating this neuron, they were able to transfer sentiments in sentences. \citet{xu2020variational} used unsupervised representation learning to separate style and content on a mixed corpus of unspecified styles. They were able to isolate a latent dimension responsible for sentiment and achieved satisfactory results in sentiment transfer. \citet{shen2020educating} %extended adversarial auto-encoders with a denoising objective. %, where original sentences are reconstructed from perturbed versions. 
%This denoising AAE model maps similar sentences to similar latent representations, making the boundaries of different representation clusters more distinct. 
%For sentiment transfer, they
computed a "sentiment vector" by averaging latent codes separately for 100 positive and 100 negative sentences in the development set,  then calculating the difference between them. Given a test sentence, they changed its sentiment from positive to negative or vice versa by adding or subtracting the sentiment vector.

The performance of these purely unsupervised TST methods is encouraging. However, it is important to note that these methods have been evaluated primarily on sentiment transfer tasks. %However, it remains unclear whether these methods can be applied to other stylistic properties, such as politeness of the text.
More research is needed to determine whether purely unsupervised methods can be generalized to other TST tasks.

\subsection{Using Large Language Models (LLMs)}
\label{sec:llms}
Large Language Models (LLMs) have revolutionized the field of natural language processing by generating coherent and contextually relevant text \cite[e.g.,][]{touvron2023llama,touvronLlamaOpenFoundation2023}. By learning complex patterns and structures from vast amounts of text data, LLMs inherently capture various linguistic styles and nuances. This capability is particularly beneficial for TST tasks.
%, where the objective is to manipulate certain language attributes while preserving the core content. Recent studies \cite{reif2021recipe, suzgun2022prompt, liu2024adaptive, mukherjee2024large} have demonstrated the utility of LLMs in TST.

A distinctive feature of LLMs is their ability to perform valuable tasks without fine-tuning, showcasing zero- and few-shot capabilities \cite{Liu:2023}. \citet{reif2021recipe} frame style transfer as a sentence rewriting task, enhancing LLMs' zero-shot performance for arbitrary TST by using task-related exemplars. \citet{suzgun2022prompt} proposed a reranking method to select high-quality outputs from multiple candidates generated by the LLM, thereby improving performance. 
%In similar vain,  proposed using prompt-based style classifier to guide editing of a given text to impose a given style.
\citet{luo-etal-2023-prompt} proposed the use of a prompt-based style classifier to guide the search for word-level edits of a given text.
Additionally, \citet{liu2024adaptive} introduced dynamic prompt generation to guide the language model in producing text in the desired style.

While prompt engineering is a prevalent approach \cite{brown2020language, jiang2020can}, LLMs are highly sensitive to prompts \cite{mishra2021reframing, zhu2023promptbench} and may not always guarantee optimal performance \cite{liu2024adaptive}. Furthermore, \citet{Ouyang:2022} indicated that larger LLMs are not always inherently superior in understanding user intent. \citet{mukherjee2024large} demonstrated that despite good results for prompting, finetuning the LLMs still leads to significant performance improvements.
%through extensive experiments in multilingual and multi-style settings that LLMs, when combined with supervised fine-tuning and promptings, can achieve significantly better results.

\subsection{TST Evaluation}
\label{sec:tst_eval}
TST tasks are mainly evaluated in three main dimensions: style transfer accuracy, content preservation, and fluency. Style transfer accuracy is typically measured using a trained style classifier \cite{jin-etal-2022-deep}. Previous works \cite{mukherjee2023leveraging, hu2022text, jin-etal-2022-deep} evaluated content preservation using the BLEU score \cite{papineni-etal-2002-bleu} and embedding similarity \cite{rahutomo2012semantic} against the input sentences. Sentence-BERT \cite{reimers2019sentence} or a similar embedding representation is used, and then cosine similarity is used for embedding similarity. The perplexity of pre-trained language models, such as GPT-2 \cite{radford2019language}, is used to estimate fluency.

\section{Applications} 
\label{applications}

We now turn to the core of our paper, i.e. the overview of TST applications.
We grouped the applications into four broad areas:
user privacy and security (Section~\ref{sec:privacy}), 
creating personalized texts (Section~\ref{sec:personalized}), dialog response generation (Section~\ref{sec:dialogue}), TST used in other NLP tasks (Section~\ref{sec:other-nlp-use}) and TST for data augmentation (Section~\ref{sec:tst_dataaug}).
We provide a high-level overview of TST applications 
% in Figure \ref{fig:tstapp} 
and examples of these applications in Table \ref{tbl:TSTapplicationexamples}.
% For a general understanding of the text transfer applications, please refer to Figure \ref{fig:tstapp}. To gain a better understanding, you can refer to Table \ref{tbl:TSTapplicationexamples} which contains specific examples of various TST applications.
%The section is structured as follows: in Section \ref{realworldapplications}, we describe application areas on which general TST can be applied, and in Section \ref{subtasksofTST}, we follow with an overview of sub-tasks of TST commonly explored in literature, together with their downstream applications.

%\subsection{Real-world Applications} \label{realworldapplications}

% \begin{figure}[t]
%     \centering
%     \includegraphics[width=0.40\textwidth]{tstapp.png}
%     \caption{Overview of TST applications. \OD{I told you already, this feels like a figure for a sake of figure -- what's the point of the circular arrangement? Why doesn't it reflect the grouping you've done in the text?}}
%     \label{fig:tstapp}
% \end{figure}

\begin{table*}[t]
\centering
\begin{adjustbox}{width=\textwidth}
\begin{tabular} {L{.18\textwidth}p{.5\textwidth}p{.05\textwidth}p{.55\textwidth}} %{llcl}

\toprule
& {\textbf{Original Text}} & \textrightarrow{} & {\textbf{Style Transferred Text}} \\
\toprule

\multicolumn{4}{c}{\textbf{User Privacy and Security (Section~\ref{sec:privacy}) }} \\
\midrule

Bias Correction$\colon$ & \hangindent=0.5em Each of them is physically handicapped. & \textrightarrow{} & \hangindent=0.5em Each of them has a physical disability. \\ % https://blog.ongig.com/diversity-and-inclusion/biased-language-examples/
& \hangindent=0.5em She was hysterical and overly emotional. & \textrightarrow{} & \hangindent=0.5em She expressed her emotions openly and passionately. \\

\cmidrule{2-4}

Fighting Offensive Language$\colon$ & \hangindent=0.5em He's an old geezer who doesn't understand technology. & \textrightarrow{} & \hangindent=0.5em He's an older individual who may not be familiar with the technology. \\
& \hangindent=0.5em You're being such a retard, it's not that difficult. & \textrightarrow{} & \hangindent=0.5em You're struggling with this, but it's not as simple as you thought. \\

\cmidrule{2-4}

Concealing Authorship$\colon$ & \hangindent=0.5em I suggest we revise the current strategy to improve our results. & \textrightarrow{} & \hangindent=0.5em The suggestion is to revise the current strategy to improve the results. \\

& \hangindent=0.5em Our research indicates that this approach is highly effective. & \textrightarrow{} & \hangindent=0.5em The research indicates that this approach is highly effective. \\

\midrule
\multicolumn{4}{c}{\textbf{Creating Personalized Texts (Section~\ref{sec:personalized}) }} \\
\midrule

Marketing and Advertisement$\colon$ & \hangindent=0.5em Our company has been in business for 10 years. & \textrightarrow{} & \hangindent=0.5em With a decade of excellence, our established company brings unparalleled expertise and experience to cater to your needs. \\
& \hangindent=0.5em Enjoy the convenience of online shopping with our user-friendly website. & \textrightarrow{} & \hangindent=0.5em Experience seamless and hassle-free online shopping with our intuitive and user-friendly website! \\
\cmidrule{2-4}

Text Simplification$\colon$ & \hangindent=0.5em The prevalence of cardiovascular diseases is higher in individuals with a family history of hypertension and dyslipidemia. & \textrightarrow{} & \hangindent=0.5em People with a family history of high blood pressure and high cholesterol are more likely to have heart problems. \\
& \hangindent=0.5em The research utilized a cross-sectional design and collected data through self-report questionnaires and objective measurements to examine the relationship between sleep duration and mental health outcomes. & \textrightarrow{} & \hangindent=0.5em The study used surveys and tests to look at how sleep duration is related to mental health in a large group of people. \\

\midrule

\multicolumn{4}{c}{\textbf{Dialog Response Generation (Section~\ref{sec:dialogue}) }} \\
\midrule

Stylized Response Generation$\colon$ & \textcolor{white}{.} \newline \textbf{User$\colon$} Hello there.&& \\

& \hangindent=0.5em \textbf{Chatbot$\colon$} What can I help you with? & \textrightarrow{} & \hangindent=0.5em \textbf{Chatbot$\colon$} Hello! I’m here to assist you with anything you need. How may I help you today? \\

& \multicolumn{3}{l}{\hangindent=0.5cm \textbf{User$\colon$} I need help with my account.} \\

& \hangindent=0.5em \textbf{Chatbot$\colon$} What's the problem? & \textrightarrow{} & \hangindent=0.5em \textbf{Chatbot$\colon$} Of course, I would be happy to assist you with your account. Could you please provide me with some more details about the issue you are experiencing? \\
\midrule

\multicolumn{4}{c}{\textbf{Use as Part of Other NLP Tasks (Section~\ref{sec:other-nlp-use}) }} \\
\midrule

Stylized Machine Translation$\colon$ & \hangindent=0.5em Mach endlich die Tür zu! \emph{(Close the door already!)} & \textrightarrow{} & \hangindent=0.5em Could you please close the door? \\
& \hangindent=0.5em Das geht dich nichts an! \emph{(That's none of your business!)} & \textrightarrow{} & \hangindent=0.5em I'd rather not share that information. \\
\cmidrule{2-4}

Stylized image captioning:&A man and a boy lying in the snow& \textrightarrow{} &
Father and son playing in the snow, enjoying the winter.\\
&Front of an old building with a sign & \textrightarrow{} &
A photo of an ugly building with a stupid sign out front.\\
\bottomrule

\end{tabular}
\end{adjustbox}

\caption{Examples of TST Applications.
%\OD{do we want examples for 3.4?}\ML{Not sure how to do that for data augmentation}
}
\label{tbl:TSTapplicationexamples}

\end{table*}

% \paragraph{Language Correction}
\subsection{User Privacy and Security}
\label{sec:privacy}

TST has several applications in the area of user privacy and security. 
It can be used to correct undesirable social biases reflected in text, to protect users' mental health by combating offensive language, and to safeguard users' privacy by disguising their writing style while still conveying the intended message.
% \OD{this is not a good intro into this section -- you should somehow summarize or abstract over all the sub-approaches here, but you basically only speak about concealing authorship (which then gets even more confusing when you go on about bias correction)}

\paragraph{Bias Correction} 
Narratives and news texts sometimes reflect social biases and stereotypes, like the traditional gender roles %of women being passive and submissive
\cite{lakoff1973language, fiske1993controlling, fast2016shirtless}.
The TST task of \textit{controllable text revision} constructs methods for rephrasing a text to a targeted style that can also include reframing its content. %The biases are corrected by altering and equalizing the way events are described. 
This leads to the development of automatic revision tools that have already proved useful in reshaping gender roles portrayed in the media~\cite{clark2018creative}.

\citet{reddy-knight-2016-obfuscating} used rule-based lexical substitution to obfuscate gender in social media and review text. 
\citet{ma2020powertransformer} proposed a controllable debiasing that aims to rewrite a given text in order to correct the implicit and potentially undesirable bias in character portrayals. 
\citet{ma2020powertransformer} proposed a controllable debiasing that aims to rewrite a given text in order to correct the implicit and potentially undesirable bias in character portrayals. They masked verbs associated with power or agency in the input text using a dictionary~\cite{sap2017connotation} and regenerated them using a GPT-based transformer. The same dictionary is also used during decoding to increase the probability of more neutral verbs.
%They also introduced the concept of PowerTransformer which unbiases text through the lens of connotation frames \cite{sap2017connotation}.\OD{unclear what it means, try to rephrase not using their own invented terms}
%that encodes pragmatic knowledge of implied power dynamics with respect to verb predicates 
\citet{pryzant2020automatically} proposed an approach to automatically convert subjective text into a neutral point of view.

% \OD{what/why is a text “inappropriately subjective”?}

\paragraph{Fighting Offensive Language}
\label{app:fight_offensive}
Even though social media platforms have undeniably positive impacts on facilitating communication, it is often the hub of offensive and abusive language use which may lead to cyberbullying~\cite{beran2005cyber, munro2011protection}. To solve this issue, many recent studies have focused on developing machine learning models for detecting hate speech \cite{davidson2017automated, xiang2012detecting, djuric2015hate, waseem2016hateful, chen2012detecting, founta2019unified} and offensive language using various text classification methods \cite{xiang2012detecting, warner2012detecting, kwok2013locate, burnap2015cyber, nobata2016abusive, davidson2017automated, founta2019unified}. 
However, these techniques only detect the problem, but do not provide actionable tools to mitigate it~\cite{janiszewski2021time}. 
Such tools could automatically suggest transforming hateful sentences into non-hateful ones, which is essentially a TST task.
%Recently, efforts have been made on the use of non-parallel data for style/content transfer \cite{shen2017style, melnyk2017improved, fu2018style} and machine translation \cite{lample2017unsupervised, artetxe2017unsupervised}. \citet{shen2017style, fu2018style} and \citet{DBLP:conf/emnlp/LampleOCDR18} used adversarial classifiers as a way to force the decoder to transfer the encoded source sentence to a different style/language.

\Citet{santos2018fighting} proposed an extension of a basic encoder-decoder architecture by including a collaborative classifier to deal with abusive language redaction. 
They also suggest using TST to prevent online conversations from derailing by suggesting a more polite alternative to the offensive message a user is about to post.
\citet{logacheva-etal-2022-paradetox} and \citet{10.1007/978-3-031-08754-7_7} collected parallel corpora of toxic and non-toxic texts and used them to fine-tune transformer-based language models.
The multi-lingual aspect of constructing detoxification tools was tackled by \citet{dementieva-etal-2023-exploring} where a joint supervised learning of translation and detoxification was considered.
An unsupervised style transfer technique for redacting offensive comments in social media was proposed in \cite{tran2020towards}.
The approach consists of  a retrieve-generate-edit pipeline that edits text in a word-restricted manner while maintaining a high level of fluency and preserving the original text content.

\paragraph{Concealing Authorship}
Text anonymisation is often required when creating public datasets or sharing data with third parties. 
It is also useful for countering the problematic task of author profiling, which aims to extract the identity of the author of the text (e.g. whistleblowers, dissidents), including privacy-invading characteristics such as gender and age~\cite{schler2006effects,10.1145/2382448.2382450}. 
TST methods can help protect user privacy by modifying text to disguise the true identity of users~\cite{reddy2016obfuscating}. 

\citet{grondahl2020effective} proposed a combinatorial paraphrasing method to hide authorship. The approach applies multiple linguistic rules to paraphrases produced by a neural method to construct a set of candidate paraphrases, from which a final text is later selected by maximising the distance to the style of the original author.
\citet{Jones_Nurse_Li_2022} found that pre-trained language models can successfully transfer the style of a given author to fool authorship detection techniques. %\OD{doesn't make sense, how do you transfer the style “to detection”?}\ML{Hopefully now makes sense xD} 
\citet{zhai-etal-2022-adversarial} proposed adversarial training that takes into account deobfuscation techniques that attempt to guess authorship. \citet{fisher-etal-2024-jamdec} addressed the problem of lack of supervised data in many domains and proposed constrained decoding, an inference-time technique applied over small language models to conceal authorship. %\OD{how is this applied?}\ML{
The method first extracts keywords from a given text and uses them as lexical constraints in a constrained diverse beam search to generate a faithful set of candidate texts. The final output is selected by automatic metrics of coherence and grammatical correctness.

\subsection{Creating Personalized Texts}
\label{sec:personalized}

TST has the potential to be used for creating personalized texts, where users can tailor the writing style of a text to suit their preferences or needs. This can have applications in various domains, such as content creation, marketing, and communication, where personalized texts can enhance user engagement, user experience, and customer satisfaction.

\paragraph{Marketing and Advertisement}
Multiple studies reveal that the use of convincing text creates an impact on people's behavior \cite{chambliss1996adults,kaptein2012adaptive}. 
This impact depends not only on the information contained in a text, but also on the way it is presented~\cite{johnstone1989linguistic,muehlenhaus2012if,darani2014persuasive}.
Insights gained from these studies have been applied to improve the effectiveness of marketing and advertising strategies  and TST techniques can be used to automatically convert a text into an effective style.
Recent studies have explored this opportunity and developed  style transfer strategies based on a user profile \cite{kaptein2015personalizing}. \citet{jin2020hooks} proposed a use-case for using these TST methods to make news headlines more attractive. 
%\MLdel{\citet{li2020stylecontent} then proposed a disentanglement-based model\OD{explain/link} to generate attractive headlines for Chinese news.}
On the other hand, \citet{li2020stylecontent} generated news headlines in a user-independent attractive style using a special encoder-decoder model combined with headline prototype retrieval.
\paragraph{Text Simplification}

%The \textit{curse of knowledge} \cite{kennedy1995debiasing} can restrict communication between experts and laymen. 
TST can be applied to create a bridge between an expert's language and a layman's understanding of it \cite{kennedy1995debiasing,tan2017internet}. The underlying notion is style transfer which aims to improve the readability of a text by reducing its complexity, for example by explaining complex terminology with simple phrases~\cite{saggion2017automatic}. 
A very widely studied use case for these methods is the simplification of encyclopedia entries to facilitate broader access to knowledge.
Research on this application has included methods for building parallel corpora~\cite{hwang-etal-2015-aligning,kajiwara2016building} and applications of neural techniques~\cite{nisioi2017exploring}.
More recently, \citet{cao2020expertise} have been working on the simplification of medical texts, focusing not only on simplifying the professional language but also on raising the expertise level of a layman's descriptions. 
To this end, they have collected a new medical dataset and experimented with two specialized text simplification and three TST approaches.%}\OD{can you also say how these people are doing it?}

%Previous methods were developed in a \cite{kajiwara2016building,nisioi2017exploring,saggion2017automatic,nishida2019multi,cao2020expertise} have studied interesting real-world TST applications in which the texts are transferred between expert and layman styles.\OD{this is really vague, it would be good to show this in more detail and explain their differences, I assume they're not all the same?} \ml{They are the same, expect for the last one...} These applications concentrate not only on simplifying the professional language but also on raising the expertise level of a layman's descriptions \cite{cao2020expertise}.

\paragraph{Writing assistants}
TST methods can also be incorporated into automatic writing assistants, which help to refine human-written texts to make them more suitable for their purpose~\cite{10.1145/3613904.3642697}.
Writing assistants can make a complaint sound more objective, a written request more polite, or business communication more friendly or professional, depending on the context~\cite{jin2022deep}.
Writing assistants also provide the functionality of the text simplification methods mentioned above, making texts written by experts more suitable for a wider audience.
Another important aspect that has recently been tackled is preserving the user's writing style while suggesting corrections~\cite{yeh2024ghostwriter}.

\paragraph{Generating similes} Generating similes amounts to finding an effective mapping of properties between two concepts. It could help in downstream applications such as creative writing assistance and literary or poetic content creation.
While most of the prior computational work has focused on simile detection \cite{niculae2014brighter, mpouli2017annotating, qadir2015learning, qadir2016automatically, zeng2020neural, liu2018neural}, there are works specifically aimed at simile generation:  \citet{chakrabarty2020generating} propose a method for automatically constructing a parallel corpus by transforming a large number of similes collected from Reddit to their literal counterpart using structured commonsense knowledge.
\citet{yang-etal-2023-fantastic} collected a simile dataset for Chinese by automatically extracting them from a corpus of student compositions. They also proposed an encoder-decoder architecture with a retrieval module for vehicle search.
\citet{he-etal-2023-hauser} proposed an automatic evaluation system for simile generation based on five criteria: relevance, logical consistency, sentiment consistency, creativity and informativeness.
% \OD{More on simile generation: \url{https://aclanthology.org/2023.acl-long.702}, \url{https://aclanthology.org/2023.acl-long.28.pdf}}
%\ODdel{They also propose to finetune a trained sequence-to-sequence model, BART \cite{lewis2019bart}, on the literal-simile pairs to gain generalizability in order to generate novel similes from a given literal sentence.}
%\OD{not sure why so much detail on this single system (when it doesn't seem special or exemplary in any way), feels a bit weird}

\paragraph{Humor Generation} 
% Understanding and generating humor is one of the goals of natural language understanding \cite{taylor2004computationally, mihalcea2005making, purandare2006humor, hempelmann2008computational}, including distinguishing between jokes \cite{weller2019humor} and generating humorous text \cite{ he2019pun, luo2019pun}. The rise in popularity of humor generation has led to a SemEval task of predicting and understanding the level of humor \cite{hossain2019president}. The work of \citet{hossain2019president} is also helpful in creating systems that can automatically generate humor.
% \OD{I kinda tried to put the similes in contact with TST, but it's even harder with this paragraph -- unless you can make a reasonable connection to TST, drop it}

Whereas understanding humor is already a challenging task that requires both natural language understanding and common sense knowledge~\cite{taylor2004computationally, mihalcea2005making, purandare2006humor, hempelmann2008computational}, generating humorous texts that match human-created one is even more difficult~\cite{ he2019pun, luo2019pun}.
Although humour generation is often realised as a natural language generation task, e.g. by completing a given text with a pun~\cite{he2019pun}, TST techniques can be used to convert existing texts into humorous ones.
For example, \citet{weller-etal-2020-humor} used TST to modify existing news headlines and demonstrated in an A/B experiment that they were just as humorous as those edited by humans.

\paragraph{Empathetic Rewriting}
As empathetic interactions are strongly correlated with symptom improvement in mental health care~\cite{norcross2002psychotherapy,elliott2018therapist}, several studies have explored the possibility of using TST methods to rewrite human or computer-generated responses to make them more empathetic.
\citet{sharma2021towards} used reinforcement learning to train GPT2-based language models that performs sentence-level edits to make the response of mental care provider more empathetic. 
The resulting method was further refined through better hyperparameter search and improved training data quality, and used in a controlled trial where the system was shown to increase conversational empathy by almost 20\%~\cite{sharma2023human}.
%\MLdel{A rewriting method that can provide additional educational feedback in the form of rewriting templates for trainee mental health workers has been proposed by \citet{perez2023verve}. They constructed a framework that first uses attention scores to mask style-relevant words, creating a template that is later filled in with a language model.}
A rewriting method that can provide additional educational feedback to trainee mental health workers has been proposed by
\citet{perez2023verve}.
They constructed a framework that first uses attention scores to mask style-relevant words in the input sentence, and then regenerates them with target-style words.
The result is a rewriting template containing a rewritten and original sentence with highlighted (masked) replaced words, providing interpretable feedback to the user.
%\OD{I don't really understand much of this, can you rephrase it? }
%\ML{better?}

\subsection{Dialog Response Generation}
\label{sec:dialogue}

Most prior research on dialogue response generation is based on grammatically correct and contextually relevant responses \cite{ritter2011data, chen2017survey}. However, these syntactically coherent responses did not always guarantee an engaging and attractive chatbot. 
\citet{kim2019comparing} carried out a study that showcased the impact of a chatbot's conversational style on users. It was noticed that when a chatbot possessed a  language style  that matched some basic human personality traits (optimistic, humorous), it scored higher in user satisfaction and had a higher average number of interaction rounds \cite{song2019generating}. %For example, a chatbot that would be able to recommend products in accordance with the persona of the customer and their profile may adopt a more effective conversational style, while the same chatbot may switch to a polite conversational style when addressing customer complaints.
These insights can be used to build a more effective chatbot, which could choose its response style according to the customer's persona, profile and the context of the conversation, for example using a more polite conversational style when dealing with customer complaints.

Nevertheless, stylized response generation faces difficulties in generating coherent responses with a particular style, especially when the target style is only embedded in unpaired texts that cannot be used directly to train the dialogue model \cite{zheng2021stylized,silva-etal-2022-polite}. 
This led to the construction of stylized response generation methods inspired by unsupervised TST approaches, but also opened up the possibility of using TST methods directly by applying them to already generated responses to impose on them the desired stylistic properties.

\paragraph{Stylized dialogue response generation}
\citet{niu2018polite} used a supervised style classification dataset to generate stylised responses without parallel data, but with a trained style text classifier to train a language model in a weakly supervised manner.
\citet{gao-etal-2019-structuring} investigated the generation of stylised responses from non-stylised dialogue data and non-dialogue text in a given style. They proposed a special neural architecture consisting of a sequence-to-sequence model for dialogue data and two autoencoders: one for stylised text and one for dialogue responses from parallel data. Both autoencoders and the sequence-to-sequence model share the same decoder and are trained together with multi-task learning, using specialised regularisers to align their latent spaces.
\citet{zheng2021stylized} also explored this type of training data and proposed an inverse dialog model to generate possible preceding dialogue contexts for non-dialogue data. Such generated stylized pseudo-dialogs are then used to train dialog model with a joint training procedure. 
A similar, but simpler approach was applied by~\citet{mukherjee-etal-2023-polite} who applied a TST model to convert dialogue responses in the training set to the desired style and later train a chatbot. 

%\citet{wang2017steering} proposed a simple training and decoding method to produce dialog responses in a specific style. Their selective sampling-based decoding methods bias the generation process with specific language style restrictions.\OD{again, you're using their keywords but I have no idea what they're doing. also, why are you breaking chronological order here?}

\citet{silva-etal-2022-polite} explored stylized task-oriented dialog agents in a transfer learning scenario across different domains, and found that direct stylized response generation consistently produced lower quality results than the two-step approach involving standard response generation followed by TST. 
Rewriting dialog responses was also explored by \citet{sun-etal-2022-stylized} in the context of stylized knowledge-grounded dialogue generation. 
They proposed using a neural system to rewrite generated dialog responses to positive/negative/polite styles by tagging style-related words and replace them with words associated with the desired style while taking into account the information from the knowledge graph. %The comparison with direct stylized response generation baselines also

% \OD{Why are the following two separate from stylized response generation? Isn't that the same thing, just different kinds of styles?}
% \OD{Also, you had more than just this in your polite chatbot paper, no?}

%\OD{maybe drop everything except the last one in this paragraph, cause it's mostly just about persona? not sure}\ML{In this longer version, I'd leave it as is.}
\paragraph{Persona-based dialogue response generation}
A particular type of style transfer studied in the dialogue community is related to \emph{persona-based dialogue systems}, where the generated response should reflect the style and personal characteristics of a selected dialog agent's persona.
\citet{DBLP:conf/acl/LiGBSGD16} encoded personas of individuals in contextualized embeddings that helped in capturing the background information and style to maintain consistency in the generated responses. The persona embedding is passed to the model at each generation step.
\citet{su19b_interspeech} followed this approach, but used generative adversarial networks and sequence-to-sequence reconstruction loss to improve both coherence and style transfer.
\citet{DBLP:conf/aaai/ZhengZHM20} also encoded personas in additional embeddings, but developed a special attention routing algorithm to dynamically reweight the influence of persona style and dialogue context during generation.
\citet{song-etal-2020-generate} used persona embeddings to predict the masking of words in the generated response and then rewrite it to a more personality-consistent one.

\citet{luan-etal-2017-multi} focused on using both dialogue data and non-dialogue texts written by personas to train the neural response generation model through multi-task learning. 
\citet{song-etal-2021-bob} likewise used two types of training data, but the non-dialogue text corpus was used to improve the consistency of the generated responses rather than to enable style transfer. 
\citet{cao-etal-2022-model} proposed a model-agnostic data manipulation strategy, including the transfer of persona features between sentences, to train persona-based dialogues as standard sequence-to-sequence models.
Finally, \citet{xu-etal-2023-towards-zero} pre-trained a persona augmentation generation model to enable zero-shot persona transfer.

\paragraph{Joint persona-based and stylized dialogue response generation}
A special case of joint persona-based and stylized generation was covered by \citet{firdaus2022being}, who focused on generating polite, personalized dialogue responses in accordance with the user's profile and consistent with their conversation history. 
Based on the user's profile, they created human annotated politeness templates and then trained an architecture with two successive decoding steps: text generation incorporating the persona and dialogue context, followed by text refinement to incorporate style (politeness).

\subsection{Use as Part of Other NLG Tasks}
\label{sec:other-nlp-use}

TST integrates with various NLP tasks, such as sentiment analysis, machine translation, and %dialogue 
data-to-text generation. TST can be used to manipulate the style of the generated text according to specific requirements, e.g.\ to translate a formal text to an informal text, or to generate text with a specific sentiment or emotion. This can improve the versatility and adaptability of NLP models in generating texts with desired style attributes.

\paragraph{Stylized machine translation}
It is desirable to have additional control of the style of the translated text in the machine translation domain. Some of the commonly used styles for TST in machine translation are politeness \cite{sennrich2016controlling} and formality \cite{niu2017study,Wu2020ADF}. 

\citet{sennrich2016controlling} used an additional formality token at the input to the neural decoder to control the use of honorific terms.
\citet{niu2017study} proposed a re-ranking strategy for phrase-based machine translation that also takes the expected level of formality as an additional input.
\citet{Wu2020ADF} developed a dataset for training machine translation systems that simultaneously transfers the style from informal Chinese to formal English.
To mitigate the problem of lack of parallel data for stylised translation for other styles and language pairs, \citet{wu2021improving} proposed using a neural translation model together with a TST model and training them simultaneously with bidirectional knowledge transfer.
Another method to address this issue was proposed by \citet{zou2024curriculum}, who developed a curriculum pre-training strategy with four learning tasks of increasing difficulty: 
standard masked language modelling (MLM), MLM between sentences in different styles, target generation based on masked target concatenated with source, and MLM between source-target translation pairs with masked style tokens.%\OD{maybe say what the tasks are?}
% \OD{feels like it would be nice to have a bit more here -- is this the only work in stylized MT? (I'm assuming the Sennrich and Niu papers aren't stylized translation? Or if they are, can you say anything more about them?)}

\paragraph{Stylized image captioning}
Image captioning is a widely studied problem in the computer vision community, which aims at factually describing image content.
However, it has been found that generating captions in a less neutral style makes them more attractive and enables new applications~\cite{mathews2016senticap,gan2017stylenet}.
Many stylised image captioning approaches build complex multimodal architectures, often inspired by TST research, but some works use TST more directly.
\citet{10.1145/3503161.3548295} proposed a framework that first extracts a style representation from a stylised text-only corpus and then uses it for caption generation. The style extraction is performed by adopting a TST method for disentangled representation~\cite{john-etal-2019-disentangled}.
\citet{wu2022learning} proposed a multi-pass decoding process using the TST module as the final processing stage. 

\paragraph{Stylistic summarization}
Another natural language generation task that is sometimes considered in conjunction with TST is text summarisation. 
\citet{chawla-etal-2019-generating} used a style classifier-based score to define a reward function for reinforcement learning of a stylised summariser.
\citet{cao-wang-2021-inference} developed two inference-time strategies to control the style of generated summaries. One restricts vocabulary with word unit prediction and the other uses a style classifier to guide text generation.
\citet{li2020stylecontent} proposed an architecture that disentangles style and content from selected prototypes and then uses the style representation during summary generation.
\citet{goyal-etal-2022-hydrasum} showed that a model with a mixture-of-experts decoder can automatically learn different summary styles under standard cross-entropy loss.

\paragraph{Stylized data-to-text generation}
Data-to-text is a natural language generation task that involves generating text describing facts from structural data (e.g. RDF triples, tables)~\cite{10215344}.
\citet{10.1145/3603374} investigated stylised data-to-text generation of product specifications on e-commerce platforms, as they noticed that advertising texts on different platforms use different text styles (more or less formal). 
They experimented with different combinations of standard data-to-text models with two TST approaches: StyleT~\cite{dai-etal-2019-style} and NAST~\cite{huang-etal-2021-nast}, as well as an end-to-end solution.

\paragraph{Software engineering}
Just like the text in natural language, the program's code has a style that encompasses code formatting, naming convention and the use of particular programming language features to express a given functionality.
Some simple code stylistic features can be applied via hard-coded rules (e.g.~indentation), but others require more advanced solutions. This is why the software engineering community started exploring TST methods.
\citet{munson2022exploringcodestyletransfer} fine-tuned a neural language model for code style transfer in Python.
This was followed by \citet{Ting_Munson_Wade_Savla_Kate_Srinivas_2023} who constructed a tool using such trained models.
\cite{10172516} improved code style transfer performance by combining code generation with code retrieval from a large code database.

\subsection{TST for data augmentation}
\label{sec:tst_dataaug}
TST methods have also been applied to tasks not directly related to generating natural language in a particular style. 
They can be used to generate artificial training data to obtain better generalizability, obtain special examples to attack neural language models, and even to generate counterfactual examples to explain model predictions.

\paragraph{Robustness to variability}
Since TST methods aim to rephrase a text without altering the text content, they can be used to augment data for most non-style-related NLP tasks.
Although using style transfer methods for this purpose is popular in the computer vision community~\cite{zheng2019stada}, the number of approaches involving text is surprisingly low.
\citet{pmlr-v70-hu17e} used TST to generate training data for sentiment classification.
\citet{chen-etal-2022-style} approached low-resouce named entity recognition by using TST together with a special constrained decoding algorithm to produce only training examples with correct tagging.

\citet{wei-etal-2023-text} proposed using TST with back-translation augmentation to improve general, non-stylized machine translation systems.
It has been found that most of the parallel corpora used to train such systems are produced by human translators who have a different style to the naturally occurring input~\cite{van-der-werff-etal-2022-automatic}. Since this limits the performance gains from back-translation, \citet{wei-etal-2023-text} proposed using TST with back-translation to transfer the style of the training texts to the naturally occurring style. Experimental comparison with several variants of back-translation showed improved performance on several popular MT metrics.

\paragraph{Robustness to attacks}
%\ML{It differs from "3.1 User privacy\&security", but potentially could be moved there...}
Several techniques for attacking deep NLP models use TST techniques.
\citet{qi-etal-2021-mind} observed that stylistic features are irrelevant to many NLP tasks and can thus be used to construct efficient adversarial and backdoor attack methods. 
\citet{pan2022hidden} showed a backdoor attack based on generating texts with certain stylistic features via TST and demonstrated its effectiveness in bypassing existing defences. Similarly, \citet{10485431} proposed the use of a multistyle backdoor attack. 
\citet{kang2024hybrid} used TST to expand the search space of an optimisation algorithm that searches for the most effective adversarial example. 
Note that just as TST techniques can be used to attack models, they can also be used to defend them. For example, adding adversarial examples to training sets has been shown to improve model robustness~\cite{chen2022adversarial}.

\paragraph{Explainable AI}
Generating counterfactual explanations is a popular way of explaining predictions of black-box machine learning models~ \cite{10.61822/amcs-2024-0009}. A counterfactual example is an instance that is similar to the one for which the prediction was made, but the model's decision was different.
Among the many possible methods for counterfactual generation, 
\citet{10.1007/978-3-030-77211-6_38} suggested using TST methods for this purpose in the medical domain.
They adapted the delete-retrieve-generate TST approach~\cite{DBLP:conf/naacl/LiJHL18} to delete sets of undesirable attributes from electronic health records, retrieve them from similar healthy patient data, and generate the final counterfactual.
Their approach has been further extended to take into account concurrent treatments and the patient's medical history~\cite{WANG2023102457}.

\section{Open Problems}
\label{sec:openproblems}

% \OD{maybe this intro is a bit too long, given that you repeat the problems in the subsections? consider shortening or moving parts of this into the subsections}
Many aspects of TST still require deeper understanding, including how to learn good style representation, whether some styles are easier to represent than others, how to measure content preservation, and what is modeled in disentangled style-content text representations. The lack of understanding of these issues correlates with the applicability of TST in real-world scenarios.
We have organized our discussion of open issues in TST around three main themes: evaluation, multiple style transfer and multilingualism - each of which is described in a separate section below.
%We conclude this section by drawing lines for further research.
% TST requires a better understanding of how to disentangle text representation in a latent space into style-independent content and stylistic attributes. This lack of understanding correlates with the applicability issues of TST in real-case scenarios. For example, if a TST method for changing the formality level of a text impacts also its original content, we cannot use it for re-writing e-mails in a more formal language. In addition to that, we lack metrics for evaluating the success and the impact of incorporating TST into real-world applications. In the following paragraphs, we look into problems in TST to their current and future applications.

\paragraph{Evaluation Metrics} The existing evaluation metrics for TST are relatively limited \cite{pang2019towards, pang2018unsupervised, mir2019evaluating,myerukola-etal-2023-dont}.

A commonly used metric for \textit{transfer accuracy} (how well the target style was matched) is based on a style classifier \cite{jin2022deep}. However, the style classification task does not involve measuring the strength and appropriateness of the style in a given text, not to mention that the classifier's accuracy is not perfect. 
It should also be noted that training a style classifier requires supervised data,\footnote{Using an unlabelled corpus of stylised texts combined with a general collection of texts for negative examples might be sufficient to train a style classifier, but the classification accuracy of such an approach may vary.} which is available for a very limited number of domains. %\ML{I wonder if this is really true: if I have stylised text, I can use the Wikipedia corpus as negative examples and train a classifier... Not sure about the classification accuracy though}
% It should also be noted that training a style classifier typically requires supervised data, which is available for a very limited number of domains. Although using unlabeled corpus of stylized texts combined with some general text collection for negative examples could be enough to train some style classifier, the classification accuracy of such an approach may vary, and obtaining high-quality labeled data for TST remains a challenge for many domains.
Although unsupervised TST methods have been proposed \cite{radford2017learning,xu2020variational,shen2020educating}, they still need labeled data for evaluation, limiting their impact in real-world applications. 
Current transfer accuracy metrics also do not take into account the context of the rewritten sentence, which has a significant impact on how humans evaluate them~\cite{myerukola-etal-2023-dont}.

Moreover, a suitable metric for measuring content preservation has not been established yet. Most commonly, BLEU \cite{papineni2002bleu} is computed between transferred sentences and their source, as the n-gram overlap with the source is considered a proxy for content preservation. Cosine similarity \cite{rahutomo2012semantic} between the embeddings of the original and the transferred sentence is also sometimes used as a metric~\cite{fu2018style}.
This methodology follows the idea that the embeddings of the two sentences should be close to each other if most of the semantics is preserved. However, none of the automatic metrics are able to evaluate style transfer methods specifically in terms of content preservation in style transfer \cite{toshevska2021review}.

% \OD{I moved the following few sentences from the start of the subsection, I think they work better here. However, it probably needs some citations, maybe some of the ones you have in the first sentence in this subsection?}
The main reason behind these limitations is the entanglement between the semantic and stylistic properties of natural language. It is very difficult to separate these two properties because changing one often affects the other.
Thus, balancing the trade-off between style transfer accuracy and content preservation poses a challenge for the objective evaluation of TST tasks.
Further analysis is also needed on how well models can represent different styles and how this affects the performance of their transfer. Better evaluation methods and tools for analyzing the weaknesses of TST models would provide a better way of assessing their usefulness in different real-world applications.

\paragraph{Multiple TST} Up until now, most works explored in TST are related to binary style transfer. However, in many real-world applications, performing \textit{multiple style transfers} at the same time is required. For instance, transferring both the formality and the demographic style could make the text more appealing to a certain audience. Multiple TST is very challenging and very little explored. \citet{lampleSSDRB19} proposed techniques to do such multiple attributes style transfer together. The specification of multiple attributes makes the TST task more complex and realistic.

\paragraph{Multilinguality} Most previous works on TST have been done in English, neglecting other languages for which language-specific TST applications could be developed. For example, an interesting application of TST could be localizing a text (e.g.~using locally known named entities) after translating it into another language. 
\citet{mizukami2015linguistic} incorporated individual Japanese writing styles in a work focused on statistical machine translation. % Exploring the TST applications in different languages can improve our general understanding of this task.
\citet{garcia2021towards} and \citet{krishna2022few} aim to address the lack of non-English works in a principled way by studying multilingual TST in low-resource scenarios. \citet{garcia2021towards} leverage a large pre-trained multilingual model and fine-tune it to perform general-purpose multilingual attribute transfer. 
\citet{krishna2022few} proposed an approach for few-shot style transfer in low-resource multilingual settings. They introduced DIFFUR, a model that enhances few-shot style transfer by modeling stylistic differences between paraphrases, achieving significant improvements in formality transfer across seven languages.
Furthermore, \citet{mukherjee2024multilingual} focus on sentiment transfer in eight Indian languages, introducing dedicated datasets and evaluating various benchmark models, including both low-resource parallel and non-parallel approaches. 
%Their work highlights the significance of parallel data in TST and also demonstrates the effectiveness of non-parallel techniques in multilingual settings.
Further down the road, this line of research could lead to the development of multilingual TST applications.

% \paragraph{Leveraging Unlabeled Data} 
% The labeled data for TST is available for a very limited number of domains, hence the need for exploration of unsupervised techniques.
%There are many domains for which only unlabeled data is available. 
% Although  unsupervised TST methods have been proposed \cite{radford2017learning,xu2020variational,shen2020educating}, most of them still need labeled data for evaluation. 
% \citet{jain2019unsupervised} attempted to evaluate formality transfer by measuring the distance between the original and stylised sentence embeddings to check for content preservation. However, to measure the effectiveness of style transfer itself, they rely entirely on human evaluation.
% Another problem is that the unlabeled techniques have only been applied to English, and their effectiveness for other languages has yet to be explored.

%Another issue is that the scarce labelled data for TST is mostly available in English only. This creates an opportunity for  multilingual transfer  from English to other languages, but such techniques in the context of TST are largely unexplored.

%Few works \cite{} have achieved TST without using labeled data but then for evaluation they have used labeled test dataset. Also, using labeled data is feasible for English, but not explored for other languages. More TST work in this direction means using unlabeled data from multiple domains and multiple languages should be explored in developing applications where unlabeled or plain text may be used for achieving the desired results.

\section{Future Directions}
\label{sec:future}

Other potential future directions for TST research include:
\begin{itemize}
    \item \textbf{Document-level TST} Existing TST techniques are mostly operating on the sentence level. Document-level TST can eventually help to transfer the style of full books or Wikipedia articles or to simplify medical and law documents, keeping the transfer consistent across the document, similar to today's document-level machine translation \cite{bao2021g}. An interesting approach to document-level TST could be incorporating contextual TST \cite{cheng2020contextual}.
    \item \textbf{Dynamically-tailored content} TST built upon smaller models with lower latency \cite{fournier2021practical} could be used for generating dynamically-tailored content in real-time, e.g., for targeting automatically generated reports or news articles to a particular audience.
    \item \textbf{Domain independence} More domain-independent TST approaches could enable using a single model for multiple applications. For example, words such as  ``\emph{delicious}'' and  ``\emph{tasty}'', convey positive sentiment in restaurant reviews but not in movie reviews where the words ``\emph{imaginative}'', ``\emph{hilarious}'', or ``\emph{dramatic}'' are more indicative. Domain shift~\cite{domianshift} thus results in feature misalignment. Only little has been explored in \citet{li2019domain} regarding domain adaptation TST.
    Their approach includes the use of domain discriminators to align features from different domains, which helps in adapting the domain shift issue and enhances the generalizability of TST models across various domains.
    \item \textbf{Combining existing TST approaches} Some of the existing TST methods could be combined or extended to find their way into new application areas without the need to collect new data sets and build new architectures. 
    For example, a text paraphrasing system could be  built as an ensemble of multiple TST systems (e.g. for formality, politeness, simplification) with the final result selected by a reranking mechanism.
    Another possibility is to combine TSTs such as politeness, formality and humor adaptation to localise content for different cultures and to adapt marketing texts to the cultural nuances of different target markets.
    %  an be used for text paraphrasing.\OD{I'm not really sure what you mean by this -- it's all paraphrasing, no?}
    \item \textbf{Improved style representation} Constructing a latent representation of text which preserves its meaning while abstracting away from its stylistic properties is not trivial \cite{lample2018multiple}, and even impossible in theory without inductive biases or other forms of supervision  \cite{locatello2019challenging}. New ways of representing text independently of its style could have a long-lasting impact on TST research.
\end{itemize}

There are even more potential TST application areas, which open new directions for future work. For instance, TST can be used for data augmentation by generating the same text with different emotions, which may be used for tasks such as automated story plot generation. A similar approach has been already explored in image style transfer \cite{zheng2019stada, jackson2019style}. TST can also be used in today's marketing-dependent and social network-based scenarios, e.g., for generating more attractive headlines. Furthermore, TST can be used for changing time- and socio-dependent elements of a text, which may help to attract new generations of readers to historical texts. The nature of paraphrasing generation \cite{madnani2010generating} shares a lot in common with TST. Many trained TST models can be borrowed for paraphrasing, such as formality transfer and simplification. By paraphrasing the texts and conversations intended for adults, TST can be also used for making the content more amenable to children. 

Also, sentiment transfer applications remain largely unexplored despite having immense potential in various fields such as customer service, social media, marketing, and news coverage. For instance, they can help provide more balanced and neutral responses to customer complaints, balance the tone and sentiment of social media posts, create engaging marketing content while preserving the core message, and provide a balanced and neutral view of events and topics in news coverage. In summary, sentiment transfer has the potential to create more positive and engaging content while providing a balanced and neutral perspective on information.

\section{Ethical Impacts of TST Applications}
\label{ethics}

% As defined by \ZK{\citet{Fieser2003-FIEE}}, \textit{ethics} is a theoretical and applied branch of philosophy which studies what is good and right, especially as it pertains to how humans ought to behave in the most general sense. Since the natural language processing domain of research is also essentially a human activity, it well belongs within the scope of ethical theory. \citet{DBLP:conf/naacl/PrabhumoyeBSB21} stated that the ethical challenges of natural language processing \ZKdel{serves as an}\ZK{are in} understanding and mitigating the bias in data and algorithm\ZK{s, mainly} by identifying objectionable content like hate speech, stereotypes and offensive language\ZKdel{ and also helps in}\ZK{. Ethics also helps with} building frameworks for better system design and data handling practices. 

% \ZKdel{Technologies can bear unintended negative repercussions} \cite{DBLP:conf/acl/HovyS16}. \ZKdel{As skillful as the TST research is in fighting hate speech, offensive language, sexist and racist language and cyberbullying, it also comes with some major negative consequences.}

%As defined by \citet{Fieser2003-FIEE}, \textit{ethics} is a branch of philosophy which studies what is good and right, especially as it pertains to how humans ought to behave in the most general sense. 
TST has to deal with the same ethical challenges as natural language processing in general, which are mostly in understanding and mitigating the bias in data and algorithms \cite{PrabhumoyeBSB21}. 
This is an especially urgent problem for TST models based on language models trained on massive uncensored datasets, as task-specifically finetuned models inherit risks of their base models  \cite{weidinger2022taxonomy}.
%Also, another risk involves that, the modern-day language models are mostly used for generative tasks that help in building representations and are based on massive, uncensored datasets. The same is then finetuned on a smaller, focused corpus that is task-specific. Unfortunately, these finetuned models inherit all the potential risks that are involved with the large foundation models \cite{weidinger2022taxonomy} reflecting the unjust, toxic, and oppressive speech present in the training data \cite{weidinger2022taxonomy}. This leads to some negative consequences like learning and projecting unknown biases that can preserve social exclusion, discrimination, and hate speech \cite{weidinger2022taxonomy}.
TST also has to follow general principles while working on topics that involve human subjects or have direct applications to humans \cite{beauchamp2001principles}. More specifically, the work on TST involves two particular ethical considerations: social impact and data privacy.

TST can be used in negative propaganda \cite{bernays2005propaganda, carey1997taking}, for instance, it can be misused by fraudulent sellers to manipulate the polarity of reviews to their benefit, utilizing sentiment transfer \cite{prabhumoye2018style}, and  for generating malicious texts which can be harmful to people in various scenarios. There have been reports of several victims who tended to be suicidal due to the stress from cyberbullying \cite{kowalski2014bullying}. The stress may further lead to detachment from family \cite{oksanen2014exposure}. This effect is particularly amplified amongst young social media users.

% \OD{maybe make the connection between TST and the bots more explicit here?}
TST can also be used to make fake text highly persuasive and tailored to specific user groups by mimicking their communication style. The given text can even be adapted to the preferences of a particular user, for example when the system is used as a social bot. 
An example cited in \citet{jin2022deep} showcases how social bots played a significant role in the 2016 United States presidential election \cite{bessi2016social, shao2018spread} by being responsible for 19\% of the total tweets. It is a prominent example of how TST may be used to manipulate major political decisions. 
% This phenomenon advocates certain ideas and  supporting campaigns which are highly sensitive issues by nature and may face the risk of being used on social bots to manipulate political views of the mass. 
As suggested by \citet{hu2022text}, social bots could also be used as an army with different persuasive styles to promote anti-social behavior such as online hate speech or cyberbullying against individuals or groups. 

Another potential major negative consequence of TST applications is data privacy. TST requires collecting huge amounts of data like ratings of reviews, which may also include sensitive information such as the gender \cite{prabhumoye2018style} and age \cite{lampleSSDRB19} of the user. To avoid such negative scenarios, \citet{DBLP:conf/ethnlp/LeidnerP17} proposed a framework that encourages organizations to build natural language processing applications with ethical considerations. This framework could also be applied to TST research wherein researchers could include an ethical review process to examine the research and development of TST applications.

% With the widespread increase in TST research, there is also a good chance of certain category of people misusing the technology which can have major effects on some real human life. Thus, \citet{beauchamp2001principles} came up with some core principles and guidelines that needs to be kept in researchers' minds while working on topics that involve human subjects or have direct application to humans work. 

% There are broadly two ethical considerations of TST applications which can pose some negative implications along with the positives. They are namely, social impacts and data privacy problems.

Regarding social impact, an important application of TST is changing malicious and abusive texts into positive ones (see Section \ref{app:fight_offensive}). TST can help to remove hate speech, offensive language, sexist and racist language, aggression, profanity, cyberbullying, harassment, trolling, and toxic language \cite{DBLP:conf/acl-alw/WaseemDWW17}. A neutral tone is something that the majority of the audience expects and demands. In the same context, a tool that can automatically detect subjectively-toned language and suggest neutrally-toned counterparts is very useful \cite{pryzant2020automatically}. The same tool can also be helpful for providing reading assistance to alert readers when subjectivity bias is hidden within any context \cite{pryzant2020automatically}.
This tool can also help readers by ensuring that the text is presented in a neutral and unbiased manner \cite{pryzant2020automatically}, which is especially important to maintain objectivity in news, encyclopedias, and other informative content.
% Efforts are being made by the day to combating toxic language scenarios. 
The automatic offensive-to-neutral language transfer can be
%a helpful intelligent assistant that can address such needs. These approaches may in turn be used on 
used for social chatbots to ensure the generated content does not contain any offensive language \cite{roller2020recipes} and to purify malicious content on social media.

% Another positive impact of TST is the use of subjective language which is the way of communication through which one expresses themselves and influences others. However, the strive for neutrality is often found in certain modes of communication like textbooks. A neutral tone is something that the majority of the audience expects and demands. In the same context, a tool that can automatically detect subjectively-toned language and suggest neutrally-toned counterparts is very useful \cite{pryzant2020automatically}. The same tool can also be helpful for providing reading assistance to alert readers when subjectivity bias is hidden within any context \cite{pryzant2020automatically}. Additionally, authors and editors also benefit from the same in terms of comprehensive reviews of new and existing content.\OD{this added section is a bit repetitive and should be integrated with the rest of the text; also, is the subjectivity classifier actually relevant for TST? I'd say that's just text classification.}

The goal of a discussion on ethics is to take into account various concerns like how a system should be built, who it is intended for, and how to assess its societal impact \cite{hovy2016social,beauchamp2001respect}. Instead of abandoning the whole idea of building such tools, one must explore the concerns and find ways to deal with them \cite{leidner2017ethical}. This should be viewed as an opportunity to increase transparency by surfacing the risks and finding the best ways to its strategy into practice.

% An important attribute of TST is to change malicious and abusive texts into positive \ZKdel{ones}. Shades of abusive language includes hate speech, offensive language, sexist and racist language, aggression, profanity, cyberbullying, harassment, trolling, and toxic language \cite{DBLP:conf/acl-alw/WaseemDWW17}.  The applications of TST are also used on social chatbots to purify bad contents that are generated. The TST application of conversion from informal to formal can also be used as a writing assistant to help make writings sound more professional. However, all of this comes with a negative side.

% While sentiment modification helps in changing negative to positive mood of a text with preserving the content, it can also sometimes be misused by companies to manipulate the polarity of reviews to their benefit. This leads to a change in a negative restaurant or product review to a positive one by fraudulent sellers. Such a technique is used a cheating method to misguide innocent customers. It can also be used to generate fake positive reviews or forge fake documents.

\section{Conclusion}

In this survey, we provided a comprehensive overview of the applications of TST. We also discussed the problems and ethical impact of TST and provided directions for future work in the domains that can make use of future TST developments.
We hope that this contribution will not only stimulate TST research by highlighting potential application perspectives and problems encountered in specific applications, but will also be useful to researchers and practitioners in other fields who are considering the use of TST in their work.
% Acknowledgments is an un-numbered section. Keep them hidden until camera-ready.

\section*{Acknowledgements}
This research was funded by the European Union  (ERC, NG-NLG,  101039303) and by Charles University projects GAUK 392221 and SVV 260698.

\bibliography{anthology,custom}
\bibliographystyle{acl_natbib}

\end{document}